# "The importance of being lazy"
## – Using lazy evaluation to process queries to HPSG grammars –

Thilo Götz and Walt Detmar Meurers*

SFB 340, Kleine Wilhelmstr. 113, 72074 Tübingen, Germany
E-mail: {tg,dm}@sfs.nphil.uni-tuebingen.de

## 1 Introduction

Linguistic theories formulated in the architecture of HPSG can be very precise and explicit since HPSG provides a formally well-defined setup. However, when querying a faithful implementation of such an explicit theory, the large data structures specified can make it hard to see the relevant aspects of the reply given by the system. Furthermore, the system spends much time applying constraints which can never fail just to be able to enumerate specific answers. In this paper we want to describe lazy evaluation as the result of an off-line compilation technique. This method of evaluation can be used to answer queries to a HPSG system so that only the relevant aspects are checked and output.

The paper is organized as follows. The next section describes three different ways to check grammaticality. In sec. 3, we introduce our lazy compilation method and compare it to a more standard compilation. We examine the theoretical properties of our approach in sec. 4 and conclude in sec. 5.

## 2 Three ways to check grammaticality

Formally speaking, a HPSG *grammar* consists of a *signature* defining the linguistic ontology and a *theory* describing the grammatical objects. A grammar G *admits* some term $\phi$ just in case G has a model that satisfies $\phi$.

**Checking Grammaticality I: Enumerating models** The simplest possibility to answer a query to a HPSG grammar is to construct the models of the grammar which satisfy the query and enumerate all possibilities. The algorithm proposed in Carpenter (1992, Chapter 15) is an example for this method. It is implemented in the type constraint part of the ALE 2.0 system (Carpenter and Penn 1994). Another computational system which can proceed in this way is TFS (Emele and Zajac 1990).[1] Since such systems give full models as answers to queries, no additional knowledge of the signature or theory is needed to interpret the answers.

While enumerating models is a correct way to check grammaticality, it has a severe disadvantage: The answers are not compact in the sense that much information which could be left underspecified is made fully explicit. This concerns in particular information that could be deduced from the signature. For example, when querying an English HPSG grammar for the lexical entry of a finite past tense verb like *walked*, a system under the simple approach enumerates solutions for every person and number assignment instead of leaving those agreement properties underspecified in the answer.[2]

**Checking Grammaticality II: Satisfying all constraints of the theory** We can avoid explicit model construction by using constraint solving techniques. This can be thought of as 'enriching' the query until all theory constraints are satisfied. For the example of *walked* above this means that no agreement information is provided in the answer if there are no grammar constraints on the agreement features of that lexical entry. Computational approaches implementing this approach are, for example, the compiler described in Götz and Meurers (1995) or the WildLIFE system (Aït-Kaci et al. 1994).

---

*The authors are listed alphabetically. The research reported here was sponsored by Teilprojekt B4 'From Constraints to Rules: Efficient Compilation of HPSG Grammars' of SFB 340 funded by the Deutsche Forschungsgemeinschaft. We would like to thank Dale Gerdemann and the anonymous reviewers for helpful comments.

[1] The TFS system in version 6.1 (1994) has several evaluation options, including an undocumented "lazy narrowing" mode, which seems to implement a lazy evaluation strategy similar to that described in this paper.

[2] Since ALE uses an open world interpretation of the type hierarchy only all appropriate attributes, but not the different subtypes will be filled in. However, standard HPSG (Pollard and Sag 1994) uses a closed world interpretation of the type hierarchy. Cf. Gerdemann and King (1993) and Meurers (1994) for some discussion.

Since these systems answer queries with descriptions satisfying both theory and query, and not with full models,[3] to interpret the replies the user needs to fill in some ontological information from the signature.

While this mode of processing queries does improve on the first approach, there still are many cases in which the system does more than necessary. Consider the lexical entry of an auxiliary verb employing an argument raising technique in the style of Hinrichs and Nakazawa (1989). Such entries are being used in most current HPSG theories for German, Dutch, French, or Italian. The idea is to specify the auxiliary to subcategorize for a verbal complement plus those arguments of that verbal complement which have not yet been saturated. As a result, the lexical entry of the auxiliary subcategorizes for an underspecified number of arguments. If the raised arguments have to obey grammar constraints, e.g., in the theory of Hinrichs and Nakazawa (1994) they are required to be non-verbal signs, this results in an infinite number of solutions to the query for such a lexical entry. The reason is that the constraints enforced by the theory need to be checked on each member of the subcategorization list, and the list is of underspecified length.

The example points out a problematic aspect of the second approach to answer queries: the system checks constraints which can never clash with the information specified in the query. To avoid making these checks, we propose to use a lazy evaluation technique.

**Checking Grammaticality III: Lazy evaluation**  The basic idea of lazy evaluation is that nodes with *more information content* should be preferred in evaluation over nodes with less information content. This suggests an on-line strategy for goal selection based on the idea of laziness. However, we would like to take a compilation approach to laziness. Instead of reordering goals on-line, we compute off-line which nodes need to checked *at all* to guarantee there is a solution. This means that our on-line proof strategy is exactly identical to the non-lazy case, but it needs to do less work.

Lazy compilation can quite easily be integrated, e.g., into the compilation method translating HPSG theories into definite clause programs described in Götz and Meurers (1995). In the next section we discuss a small HPSG example to illustrate this.

Theoretically, on the other hand, lazy evaluation changes our perspective on program semantics. Whereas the programs previously had the property of *persistence* (any term subsumed by a solution was also a solution), the compilation technique for lazy evaluation abandons this property to be able to compute more efficiently. It simply demands that if $\phi$ is a solution and there are terms more specific than $\phi$, then *some* of these more specific terms must also be solutions. Such an interpretation is only correct if we impose a well-formedness condition on our grammars. This idea is due to Aït-Kaci et al. (1993), who impose a strong syntactic restriction on their theories, which will be discussed in sec. 4.1. We replace this restriction by a weaker semantic one, which demands that the grammar has some model where every type has a non-empty denotation. From the viewpoint of the user of such a system this means that not every instantiation of an answer given by the system actually is grammatical. We will illustrate this in sec. 3.2.

## 3  A HPSG example

To illustrate the second and third method to check grammaticality introduced in the last section, we want to discuss a small HPSG example: a grammar dealing with part of the agreement paradigm of German adjectives discussed in Pollard and Sag (1994, pp. 64–67).[4]  The grammar deals with the adjectival agreement pattern shown in fig. 1.

|        | fem                | masc                 |
|--------|--------------------|----------------------|
| strong | *kleine Sorge*     | *kleiner Erfolg*     |
| weak   | *(die) kleine Sorge* | *(der) kleine Erfolg* |

Figure 1: Part of the paradigm of nominative adjectives

The signature of our example grammar is shown in fig. 2.

---

[3]Some approaches even remove information deducible from the signature to keep datastructures small (Götz 1994).

[4]The small example grammar presented here only serves to illustrate the different methods of checking grammaticality. It differs in many respect from the linguistic theory developed in Pollard and Sag (1994, pp. 88-91).

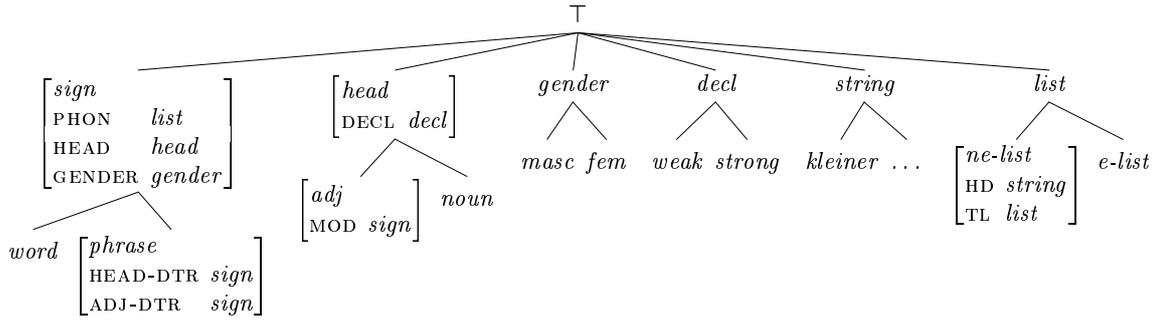

Figure 2: The signature

In fig. 3, the lexicon is defined. It contains lexical entries for the adjective *klein* (small) and for the nouns *Erfolg* (success) and *Sorge* (worry). Note that the entry for the female form of the adjective, *kleine*, is underspecified for the declension pattern.

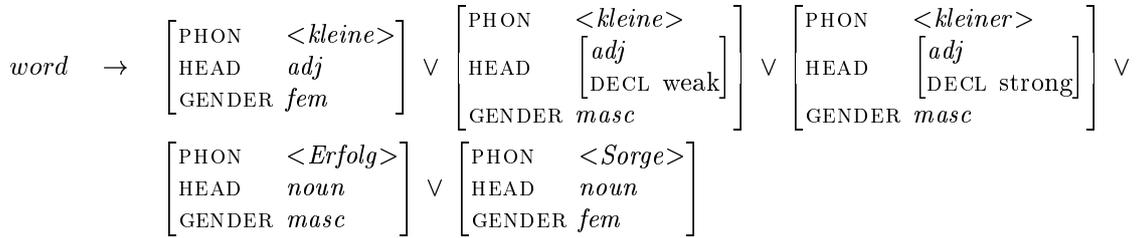

Figure 3: The word principle

Figure 4 shows a head-adjunct ID schema including the effect of the HFP and the Semantics Principle to percolate the gender (which normally is part of the content's index).

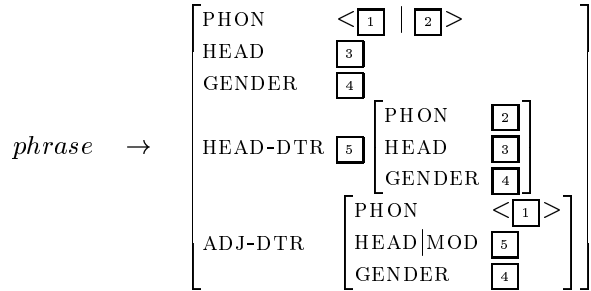

Figure 4: A simple head-adjunct ID schema

Finally, a principle is included to ensure that the declension class of the modified head is identical to that of the modifier.

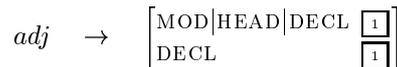

Figure 5: A principle for adjective declension

The tree in fig. 6 is an example for an agreement mismatch in a structure with two adjectives.[5] The above grammar correctly rules out this ungrammatical example, since the principle in fig. 5 enforces tags 3 and 3' in the description of *kleine* to be identical, which results in an inconsistent structure.

---
[5] The GENDER values are left out for space reasons. They are all *masc*.

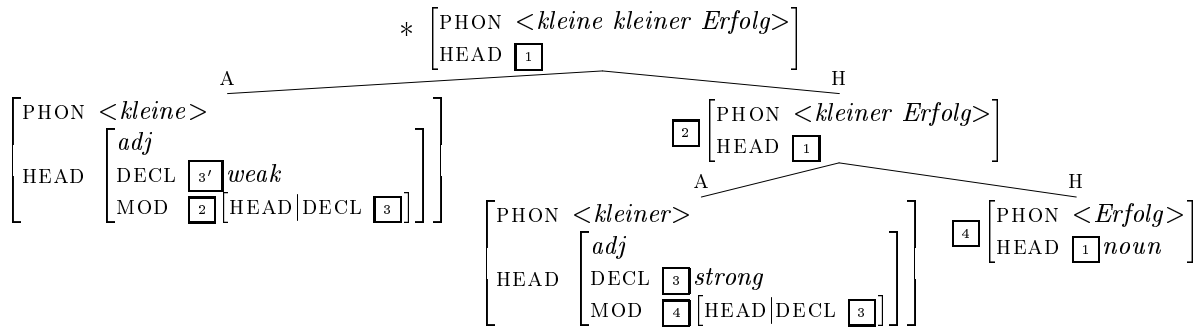

Figure 6: An example for an agreement mismatch

## 3.1 Non-lazy Compilation

In the following, we first show how the grammar defined above is compiled in a setup checking grammaticality by method II. In sec. 3.2 we then discuss how the grammar code produced by a compiler for laziness differs and how this changes processing.

A compiler, such as the one described in Götz and Meurers (1995), takes the HPSG grammar defined in the last section, determines which nodes in a structure of a certain type need to be checked, and produces code for checking these nodes. More specifically, this compiler translates constraints into clauses whose bodies are just tags that occur in the head of the clause. In the following we're interested in the question *which* nodes should be checked.

Figure 7 shows our example grammar 'in compiled form'.[6] The nodes which need to be checked are indicated with double boxes. For example, to make sure that a *word* with phonology *kleiner* is grammatical, we need to check that the *adjective* head value meets the principle for adjective declension. In fig. 7 this is marked by tag C.

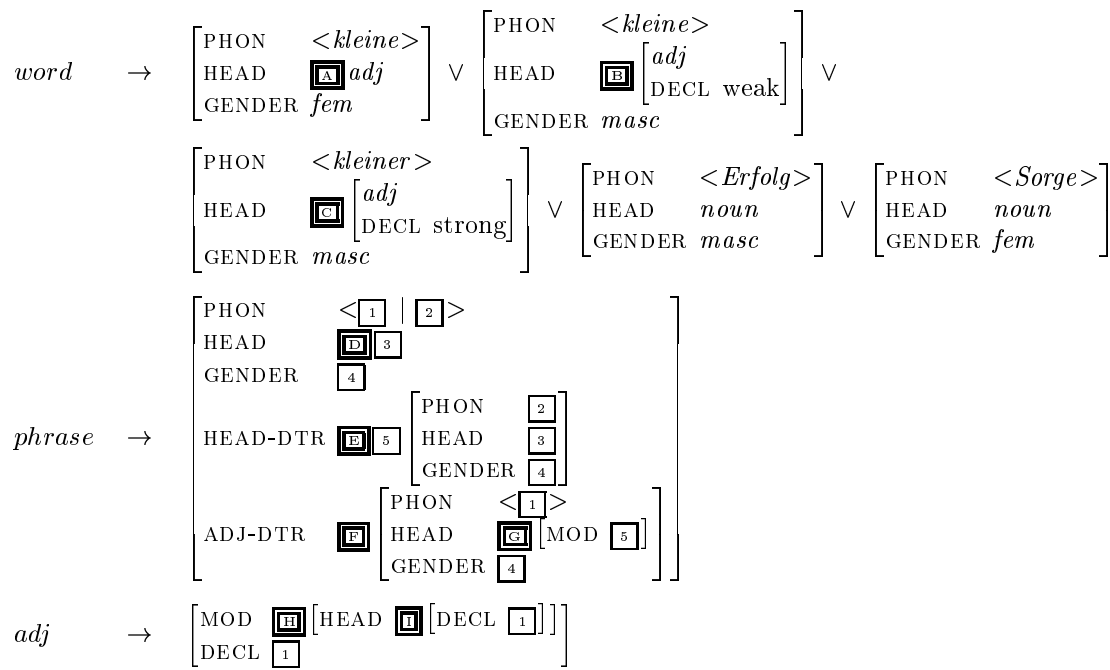

Figure 7: The compiled example grammar

---

[6]We here ignore the optimizations discussed in Götz and Meurers (1995) since they are independent of the lazy evaluation issue discussed in this paper. Briefly said, the compiled example grammar produced by the optimized version of the compiler would not include the tags D, E, and G, since those nodes will be checked when the *word* constraint is checked on node F. The same holds for I.

Now that we have a compiled grammar, let us take a look at how a query is processed. Figure 8 shows the trace of the query for a *word*, i.e., a lexical entry. In the first step, the constraint on *word* is applied. There are several disjuncts; we take the first one, the lexical entry for *kleine* (leaving a choice point behind). The disjunct chosen contains the tag A, a call to the definition of *adj*. Upon execution of that call in step 2, we unify in the definition of *adj*, which adds H, a call to *sign*, to the goal list to ensure that the sign below the MOD attribute is also a grammatical sign. Executing the call to *sign*, we again have a choice, this time between the two constrained subtypes *word* and *phrase*. So in step 3 we again choose the first disjunct of *word*, which brings us into a state where we again have to check that the *adj* at I&A satisfies the grammar - and we see that we're in an infinite loop.

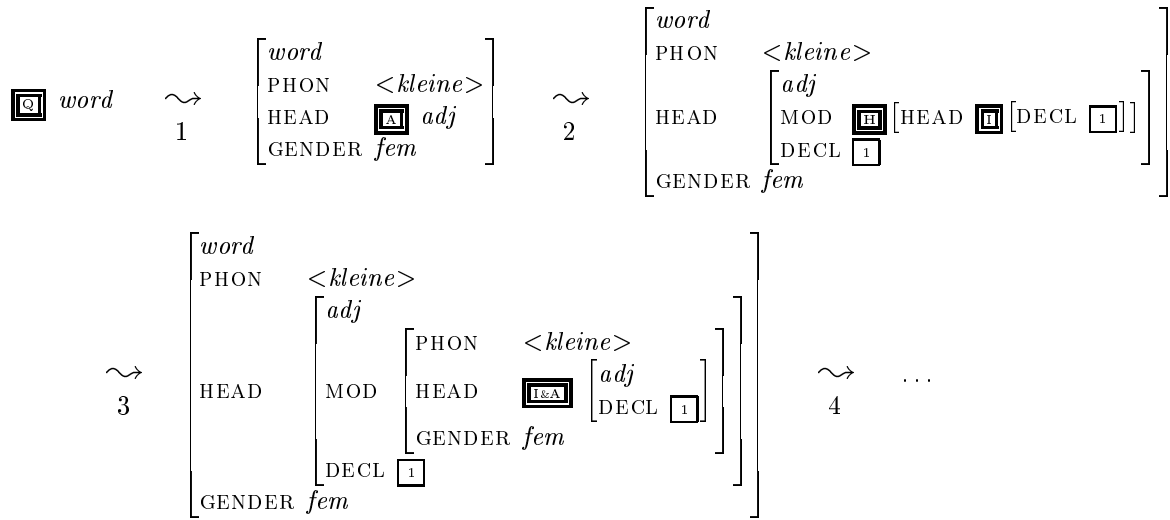

Figure 8: The trace of a query for a *word*

To obtain some answer, e.g., in step 1 we can chose another disjunct in the definition of *word* such as the lexical entry of either noun instead of the lexical entry of an adjective. There are no calls in the lexical entry of these nouns and there are none left on our goal list, so in the next state we're done. However, such a 'correct' order in which to try disjuncts needs to be specified by hand and in any case the system will still find infinitely many solutions for the above example.

## 3.2 Lazy Compilation

The lazy compilation method is almost the same, except that *no nodes without features are marked*. As mentioned in the beginning, the on-line proof strategy is identical to the non-lazy case, but since less nodes are marked, it needs to do less work. Essentially, the lazy compilation algorithm can be described as follows:

For each $t \rightarrow \Phi$ in the theory
    For each node $q$ in $\Phi$ (except the root node)
        If the type of $q$ subsumes a constrained type and at least one feature is defined on $q$
        Then mark $q$

For the example grammar in fig. 7, this means that the indices A and D go away. The intuition derives from the property we require grammars to obey for lazy compilation: *type consistency*. A grammar is type consistent iff it has a model where every type has a non-empty denotation. If there are no features defined on a node, then by type consistency, models satisfying that node exist and we don't need to search any further. We will take a closer look at this condition in sec. 4.

Returning to the example, consider the lexical entry for the feminine form of the adjective *klein* repeated in fig. 9. The HEAD value is specified to be *adj*, a constrained type. However, there is no

$$\begin{bmatrix} \text{PHON} & <kleine> \\ \text{HEAD} & \boxed{A}\,adj \\ \text{GENDER} & fem \end{bmatrix}$$

Figure 9: The compiled lexical entry of the feminine form of the adjective *klein*

feature specified for the HEAD value, and so the entry in the grammar after lazy compilation is simply as shown in fig. 10.

$$\begin{bmatrix} \text{PHON} & <kleine> \\ \text{HEAD} & adj \\ \text{GENDER} & fem \end{bmatrix}$$

Figure 10: The lazily compiled lexical entry

In the lazy approach, a query is also processed by the lazy compiler. This leads to interesting behavior: If we pose the same query as in the last section (fig. 8), namely just *word*, the system immediately comes back with the answer *word*, without further instantiating the query. This is because, by type consistency, objects of type *word* are known to exist, and no further inferences are necessary. We have to be more specific if we want to see a specific *word*. Figure 11 shows what happens if ask for a *word* with the PHON value <kleine>. With the lazily compiled grammar, we don't go into an infinite

$$\boxed{Q}\begin{bmatrix} word \\ \text{PHON} & <kleine> \end{bmatrix} \quad \underset{1}{\leadsto} \quad \begin{bmatrix} word \\ \text{PHON} & <kleine> \\ \text{HEAD} & adj \\ \text{GENDER} & fem \end{bmatrix}$$

Figure 11: The evaluation of a more specific query

loop anymore on the unspecified HEAD value. Thus, our method of lazy evaluation not only results in an efficiency increase, but actually leads to better termination properties. Of course, lazy compilation can not solve all termination problems. The problem remains for the masculine form of *klein*, whose compiled form is still as shown in fig. 12, since the HEAD value has the feature DECL specified.

$$\begin{bmatrix} \text{PHON} & <kleiner> \\ \text{HEAD} & \boxed{C}\begin{bmatrix} adj \\ \text{DECL} & strong \end{bmatrix} \\ \text{GENDER} & masc \end{bmatrix}$$

Figure 12: The lazily compiled lexical entry of the masculine form of *klein*

Reconsider the answer the system gave to the query in fig. 11. We know that our grammar contains a constraint on the type *adj* which has not been applied to the answer. In fact, precisely this constraint was at the basis of the infinite loop in non-lazy evaluation. The user therefore has to be aware of the fact that only certain *adj* objects are grammatical and that this information is not provided in the answer. The system only checks on nodes with features since those nodes are the only ones that can lead to an inconsistency.

## 4 Theoretical aspects

In this section, we will briefly look at the theoretical aspects of the proof method proposed in this paper. Specifically, we will compare our approach to the one of Aït-Kaci et al. (1993), from which we differ in two respects:

- Our basic formalism is different. We use a closed world interpretation of the type hierarchy, and we allow disjunction and negation. The basic formalism we employ therefore is the same as that used in standard HPSG. This difference has consequences for the well-formedness condition on grammars we propose as an alternative to the one given by Aït-Kaci et al. (1993).

- We compile the information about lazy evaluation off-line. The actual proof method is then very similar to SLD resolution. Aït-Kaci et al. (1993) use a more sophisticated, on-line method. Our method is essentially a simplification of theirs.

From a theoretical point of view, the most interesting aspect of a lazy evaluation method is its soundness. Since we do less work in our proofs, we need to ensure that we don't stop resolving too early. We must make sure that when our proof terminates, there are no contradictions hidden in the search space that we just didn't get to because of our laziness. The example theory in fig. 13 will illustrate that lazy evaluation is not sound in general.

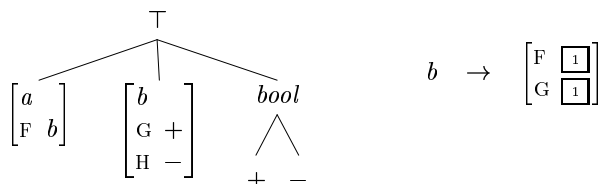

Figure 13: Unsoundness of lazy evaluation for non-type consistent grammars

Consider the query in fig. 14. Our method will say that there's nothing to prove here: There are no constraints on $a$, and the $b$ node is a terminal node and thus it does not need to be checked. However, the constraint on $b$ is clearly inconsistent. There can never be any models of this grammar with objects of type $b$ in them. By the appropriateness conditions it follows that there can not be any objects of type $a$, either. So our proof system should really come back with the answer **no**.

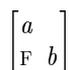

Figure 14: A query for the unsoundness example

Aït-Kaci et al. (1993) solve this problem by giving a sufficient syntactic condition for grammars that ensures soundness, i.e., by restricting the class of grammars that they can handle. Indeed, it is very hard to imagine a lazy proof system that is sound for all grammars. We will thus also restrict our attention to a proper subset of possible grammars. However, instead of using the syntactic restriction of Aït-Kaci et al. (1993), called *well-formedness*, which we suggest below to be too strong for HPSG grammars, we will use a weaker semantic one. We say that a grammar is *type consistent* iff for every type $t$, there is a model of the grammar that contains at least one object of type $t$. That is a very reasonable restriction, since one might expect the grammar writer not to introduce any types that never denote anything. One can show that our lazy resolution method is sound with respect to type consistent grammars.

## 4.1 Well-formedness vs. type consistency

The condition of type consistency is properly weaker than that of well-formedness, the syntactic condition of Aït-Kaci et al. (1993). Every grammar that is well-formed is also type consistent, but not vice versa. We conjecture that the soundness result of Aït-Kaci et al. (1993) also holds for theories that are only type consistent. However, the stronger syntactic condition has the advantage of being checkable – it is decidable if a given theory is well-formed or not. It is in general undecidable if a theory is type consistent. But note that it is also undecidable whether a given theory can be transformed into an equivalent one that meets the syntactic condition of Aït-Kaci et al. (1993). For theoretical considerations, it is still useful to use our semantic restriction, since it is the weakest possible condition for soundness of the kind of lazy evaluation that we use, i.e., it is a necessary condition. We can thus try to find weaker, checkable sufficient conditions that are more suitable for the kind of linguistic applications that we have in mind. As long as they entail type consistency, they will always guarantee soundness of lazy constraint solving.

We will now illustrate the difference between well-formedness and type consistency with two examples. The first one is trivial and shows the general idea, the second one is more practical and involves disjunction. Simplifying somewhat, the condition of well-formedness requires that for each consequent in the grammar, unfolding the type constraints for each node exactly once would not add any new information.

Suppose we have a type hierarchy of types $a$, $b$ and $c$, which are minimally ordered such that $a$ subsumes $b$ and $c$. Consider the constraint shown in fig. 15. This theory is not well-formed (unfolding

$$b \;\rightarrow\; \left[ \text{F} \; \left[ \begin{array}{c} b \\ \text{F} \; a \end{array} \right] \right]$$

Figure 15: A theory that is not well-formed

the node labeled $b$ will bump the node labeled $a$ to $b$), but it is type consistent. Moreover, the theory can not be brought into well-formed format through partial evaluation: the process will not terminate. However, one could substitute the equivalent $b \;\rightarrow\; \left[ \text{F} \; b \right]$ to obtain a well-formed theory.

A more realistic example is the junk slot encoding of the append relation (Aït-Kaci 1984). We here assume an appropriate extension of the well-formedness condition to disjunctive theories. The theory

$$append \;\rightarrow\; \left[ \begin{array}{l} \text{ARG1} \; \langle\rangle \\ \text{ARG2} \; \boxed{\text{L}} \\ \text{ARG3} \; \boxed{\text{L}} \end{array} \right] \;\vee\; \left[ \begin{array}{ll} \text{ARG1} \; <\boxed{\text{H}} \,|\, \boxed{\text{T1}}> \\ \text{ARG2} \; \boxed{\text{L}} \\ \text{ARG3} \; <\boxed{\text{H}} \,|\, \boxed{\text{T2}}> \\ \text{JUNK} & \left[ \begin{array}{ll} append \\ \text{ARG1} & \boxed{\text{T1}} \\ \text{ARG2} & \boxed{\text{L}} \\ \text{ARG3} & \boxed{\text{T2}} \end{array} \right] \end{array} \right]$$

Figure 16: The junk-slot encoding of append

in fig. 16 is not well-formed, although type consistent. Consider what happens if we try to unfold this type definition with respect to itself as shown in fig. 17. Unfortunately, the result is still not well-formed, and indeed we can not get a well-formed type constraint for *append* by unfolding or any other transformation.

We conclude that in a setup without disjunction and with open-world reasoning, like the one originally proposed by Aït-Kaci et al. (1993), well-formedness is a useful strengthening of type consistency. In a HPSG setup, using closed world reasoning and disjunction[7], well-formedness appears to be too strong.

---

[7]Note that disjunction does not increase the expressive power of a system under closed world reasoning, since disjunction can be expressed via the type hierarchy. This is different in an open world setup, where disjunction is needed to enforce a choice of subtypes.

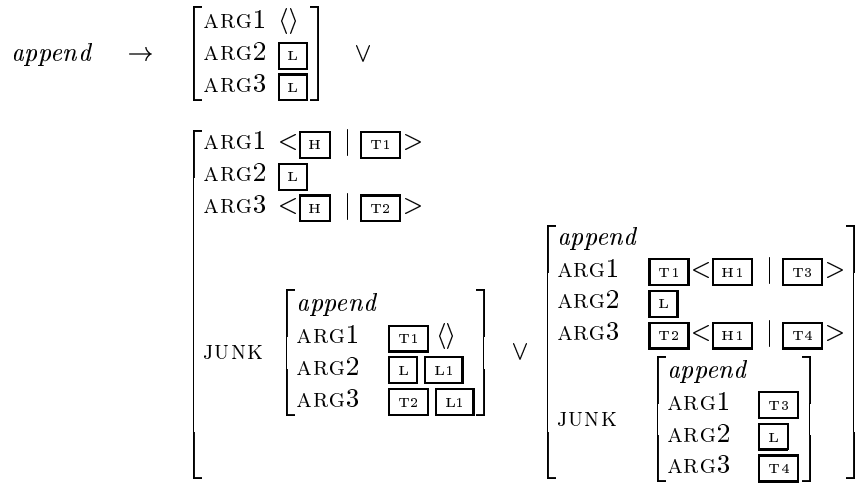

Figure 17: The junk-slot encoding of append after one unfolding step

Therefore, a more liberal syntactic restriction needs to be found. In the meantime, the grammar writer needs to ensure that our semantic condition, type consistency, is met.

## 5 Conclusion

In this paper, we have discussed three possibilities to answer queries to a HPSG grammar. We described a compiler that takes a HPSG grammar and compiles it such that standard evaluation yields a lazy strategy. Lazy evaluation in our approach therefore is not an on-line goal reordering strategy, but the result of an optimizing compiler. This removes the overhead of on-line goal reordering from processing. We showed that lazy compilation has advantages both for efficiency of processing and the termination properties of HPSG grammars.

Theoretically, we justified our approach by giving the weakest possible condition that guarantees soundness of lazy evaluation: type consistency. We argued that this simplifies the search for stronger, checkable conditions. One only needs to show that a candidate condition is stronger than type consistency; no separate soundness proof is required.

The lazy compiler described has has been fully implemented as part of the ConTroll system (Götz in prep). So far it has been tested with two complex HPSG grammars for German: one implementing the theory proposed in Pollard (1990) and the other focusing on the phenomena of aux-flip and PVP-topicalization (Hinrichs and Nakazawa 1989, 1994). Lazy evaluation for these grammars led to efficiency gains of up to 30% compared to the non-lazy approach described in Götz and Meurers (1995).

Many of the referenced papers by the authors and their colleagues are available electronically over the URL: http://www.sfs.nphil.uni-tuebingen.de/sfb/b4.